\documentclass[
]{ceurart}

\usepackage{scalerel, xparse}
\NewDocumentCommand\emojismilesweat{}{
    \scalerel*{
        \includegraphics{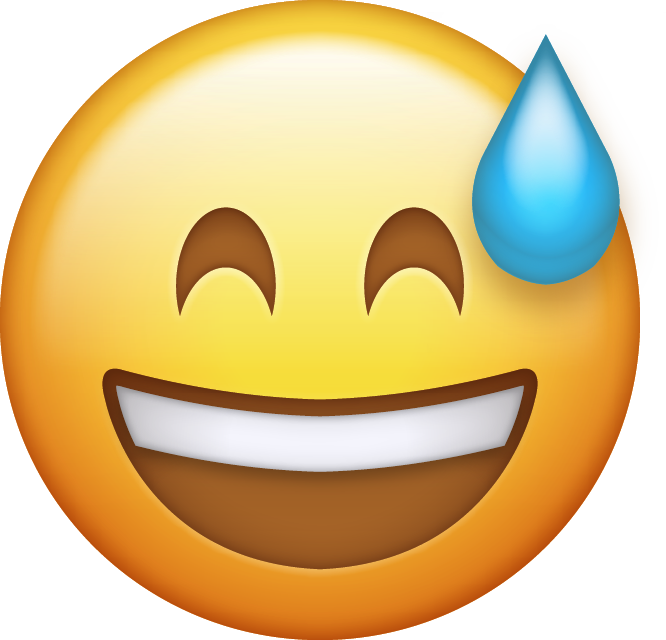}
    }{X}
}

\usepackage{scalerel, xparse}
\NewDocumentCommand\emojitongue{}{
    \scalerel*{
        \includegraphics{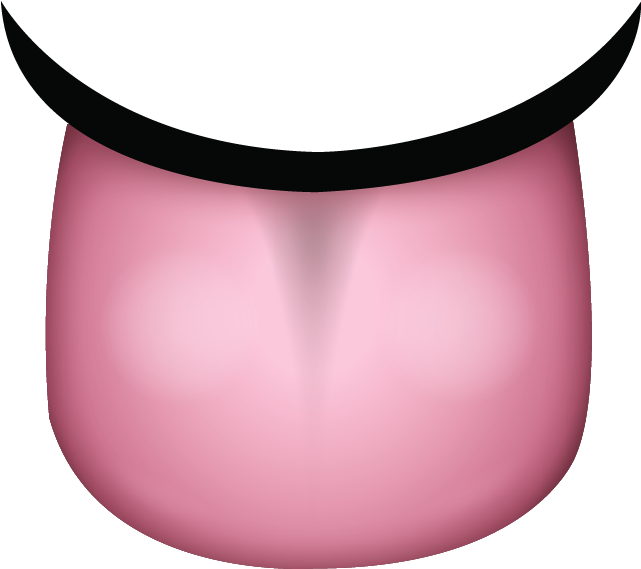}
    }{X}
}

\usepackage{scalerel, xparse}
\NewDocumentCommand\emojipanda{}{
    \scalerel*{
        \includegraphics{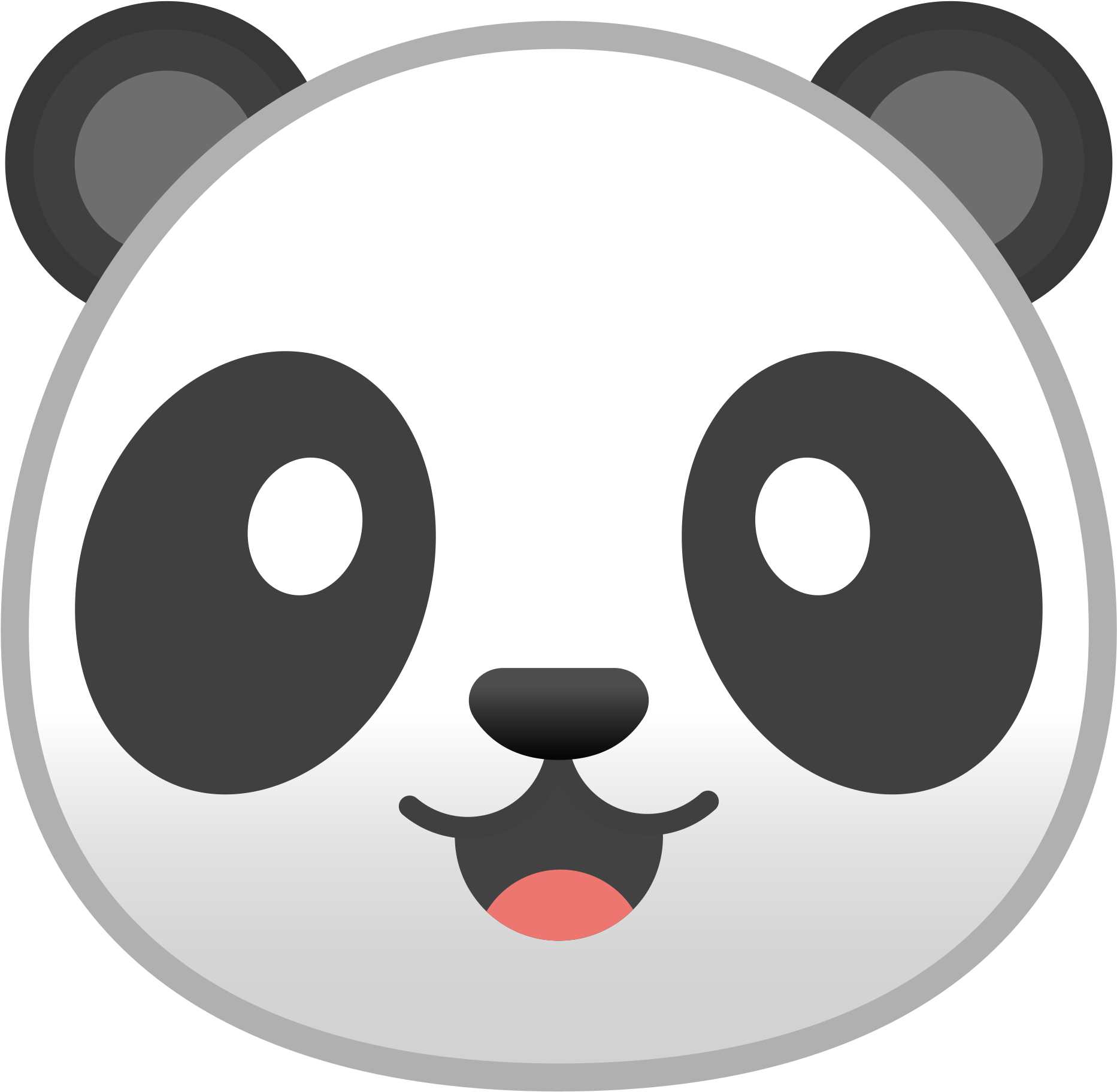}
    }{X}
}

\usepackage[justification=centering]{caption}

\begin{document}
\copyrightyear{2020}
\copyrightclause{Copyright for this paper by its authors.
  Use permitted under Creative Commons License Attribution 4.0
  International (CC BY 4.0).}

\conference{FIRE '20, Forum for Information Retrieval Evaluation, December 16--20, 2020, Hyderabad, India}

\title{Leveraging Multilingual Transformers for Hate Speech Detection}

\author{Sayar Ghosh\ Roy}[%
email=sayar.ghosh@research.iiit.ac.in,
url=https://sayarghoshroy.github.io/,
]

\author{Ujwal Narayan}[%
email=ujwal.narayan@research.iiit.ac.in,
url=https://www.ujwalnarayan.ml/,
]

\author{Tathagata Raha}[%
email=tathagata.raha@research.iiit.ac.in,
url=https://github.com/tathagata-raha/,
]

\author{Zubair Abid}[%
email=zubair.abid@research.iiit.ac.in,
url=https://zubairabid.com/,
]

\author{Vasudeva Varma}[%
email=vv@iiit.ac.in,
url=https://irel.iiit.ac.in/vasu/index.html,
]

\address[]{Information Retrieval and Extraction Lab, International Institute of Information Technology, Hyderabad, India}
\begin{abstract}
Detecting and classifying instances of hate in social media text has been a problem of interest in Natural Language Processing in the recent years. Our work leverages state of the art Transformer language models to identify hate speech in a multilingual setting. Capturing the intent of a post or a comment on social media involves careful evaluation of the language style, semantic content and additional pointers such as hashtags and emojis. In this paper, we look at the problem of identifying whether a Twitter post is hateful and offensive or not. We further discriminate the detected toxic content into one of the following three classes: (a) Hate Speech (HATE), (b) Offensive (OFFN) and (c) Profane (PRFN). With a pre-trained multilingual Transformer-based text encoder at the base, we are able to successfully identify and classify hate speech from multiple languages. On the provided testing corpora, we achieve Macro F1 scores of 90.29, 81.87 and 75.40 for English, German and Hindi respectively while performing hate speech detection and of 60.70, 53.28 and 49.74 during fine-grained classification. In our experiments, we show the efficacy of Perspective API features for hate speech classification and the effects of exploiting a multilingual training scheme. A feature selection study is provided to illustrate impacts of specific features upon the architecture's classification head. 
\end{abstract}


\maketitle

\section{Introduction}
With a rise in the number of posts made on social media, an increase in the amount of toxic content on the web is witnessed. Measures to detect such instances of toxicity is of paramount importance in today's world with regards to keeping the web a safe and healthy environment for all. Detecting hateful and offensive content in typical posts and comments found on the web is the first step towards building a system which can flag items with possible adverse effects and take steps necessary to handle such behavior.

In this paper, we look at the problem of detecting hate speech and offensive remarks within tweets. More specifically, we attempt to solve two classification problems. Firstly, we try to assign a binary label to a tweet indicating whether it is hateful and offensive (class HOF) or not (class NOT). Secondly, if the tweet belongs to class HOF, we classify it further into one of the following three possible classes: (a) HATE: Contains hate speech, (b) OFFN: Is offensive, and (c) PRFN: Contains profanities.

The language in use on the web is in a different text style as compared to day-to-day speech, formally written articles, and webpages. In order to fully comprehend the social media style of text, a model needs to have knowledge of the pragmatics of emojis and smileys, the specific context in which certain hashtags are being used, and it should be able to generalize to various domains. Also, social media text is full of acronyms, abbreviated forms of words and phrases, orthographic deviations from standard forms such as dropping of vowels from certain words, and contains instances of code mixing.

The escalation in derogatory posts on the internet has prompted certain agencies to make toxicity detection modules available for web developers as well as for the general public. A notable work in this regard is Google's Perspective API\footnote{\url{https://www.perspectiveapi.com/}} which uses machine learning models to estimate various metrics such as toxicity, insult, threat, etc., given a span of text as input. We study the usefulness of these features for hate speech detection tasks in English and German.

In recent years, utilizing Transformer-based \cite{vaswani2017attention} Language Models pre-trained with certain objectives on vast corpora \cite{devlin2018bert} has been crucial to obtaining good representations of textual semantics. In our work, we leverage the advances in language model pre-training research and apply the same to the task of hate speech detection. Lately, we have witnessed the growing popularity of multilingual language models which can work upon input text in a language independent manner. We hypothesize that such models will be effective on social media texts across a collection of languages and text styles. Our intuition is experimentally verified as we are able to obtain respectable results on the provided testing data for the two tasks in question.


\section{Related Work}

In this section, we will provide a brief overview of the variety of methods and procedures applied in attempts to solve the problem of hate speech detection. Approaches using Bag of Words (BoW) \cite{Kwok2013LocateTH} typically lead to a high number of false positives. They also suffer from data sparsity issues. In order to deal with the large number of false positives, efforts were made to better characterize and understand the nature of hate speech itself. This led to the formation of finer distinctions between the types of hate speech \cite{wang2014cursing}; in that, hate speech was further classified into ``profane" and ``offensive". Features such as N-gram graphs \cite{phdthesis} or Part of Speech features \cite{chen2012detecting} were also incorporated into the classification models leading to an observable rise in the prediction scores. 

Later approaches used better representation of words and sentences by utilizing semantic vector representations such as word2vec \cite{mikolov2013distributed} and GloVe \cite{pennington2014glove}. These approaches outshine the earlier BoW approaches as similar words are located closer together in the latent space. Thus, these continuous and dense representations replaced the earlier binary features resulting in a more effective encoding of the input data. Support Vector Machines (SVMs) with a combination of lexical and parse features have been shown to perform well for detecting hate speech as well. \cite{chen2012detecting}

The recent trends in deep learning led to better representations of sentences. With RNNs, it became possible to model larger sequences of text. Gated RNNs such as LSTMs \cite{sutskever2014sequence} and GRUs \cite{chung2014empirical} made it possible to better represent long term dependencies. This boosted classification scores, with LSTM and CNN-based models significantly outperforming character and word based N-gram models. \cite{badjatiya2017dlhate} Character based modelling with CharCNNs \cite{zhang2015character} have been applied for hate speech classification. These approaches particularly shine in cases where the offensive speech is disguised with symbols like `*', `\$' and so forth. \cite{character-abuse}

More recently, attention based approaches like Transformers \cite{vaswani2017attention} have been shown to capture contextualized embeddings for a sentence. Approaches such as BERT \cite{devlin2018bert} which have been trained on massive quantities of data allow us to generate robust and semantically rich embeddings which can then be used for downstream tasks including hate speech detection. 


There have also been a variety of open or shared tasks to encourage research and development in hate speech detection. The TRAC shared task \cite{ws-2018-trolling} on aggression identification included both English and Hindi Facebook comments. Participants had to detect abusive comments and distinguish between overtly aggressive comments and covertly aggressive comments. OffensEval (SemEval-2019 Task 6) \cite{zampieri-etal-2019-semeval} was based on the the Offensive Language Identification Dataset (OLID) containing over 14,000 tweets. This SemEval task had three subtasks: discriminating between offensive and non-offensive posts, detecting the type of offensive content in a post and identifying the target of an offensive post. At GermEval, \cite{germeval-task-2} there was a task to detect and classify hurtful, derogatory or obscene comments in the German language. Two sub-tasks were continued from their first edition, namely, a coarse-grained binary classification task and a fine-grained multi-class classification problem. As a novel sub-task, they introduced the binary classification of offensive tweets into explicit and implicit.

\section{Dataset}

\begin{table}[]
\begin{tabular}{lll}
\toprule
\textbf{Language} & \textbf{Train} & \textbf{Test} \\
\midrule
English  & 3708  & 814  \\
German   & 2373  & 526 \\
Hindi    & 2963  & 663 
\end{tabular}
\caption{Dataset Size (in number of tweets)}
\label{tab:datastats}
\end{table}

The datasets for the tasks were provided by the organizers of the HASOC '20\footnote{\url{https://competitions.codalab.org/competitions/26027}}. \cite{hasoc2020overview} The data consists of tweets from three languages: English, German and Hindi, and was annotated on two levels. The coarse annotation involved a binary classification task with the given tweet being marked as hate speech (HOF) or not (NOT). In the finer annotation, we differentiate between the types of hate speech and have four different formal classes:

\begin{table}[]
\begin{tabular}{lll}
\toprule
\textbf{Language} &\textbf{NOT} & \textbf{HOF} \\
\midrule
English  & 1852 & 1856 \\
German   & 1700 & 673 \\
Hindi    & 2116 & 847 
\end{tabular}
\caption{Training set label distribution: Task 1}
\label{tab:datacoarse}
\end{table}

\begin{enumerate}
    \item \textbf{HATE}: This class contains tweets which highlight negative attributes or deficiencies of certain groups of individuals. This class includes hateful comments towards individuals based on race, political opinion, sexual orientation, gender, social status, health condition, etc.\break
    Example: ``RT @Lubchansky: good to know rich people have always been dumb as shit https://t.co/otdmH0wquk''
    \item \textbf{OFFN}: This class contains tweets which are degrading, dehumanizing or insulting towards an individual. It encompasses cases of threatening with violent acts. \\
    Example: ``By shitting yourself and taking the backdoor out, instead of fronting up to the public.''
    \item \textbf{PRFN}: This class contains tweets with explicit content, profane words or unacceptable language in the absence of insults and abuse. This typically concerns the usage of swearwords and cursing. \\
    Example: ``@HermesCxbin turn that shit off''
    \item \textbf{NONE}: This class contains the tweets which do not fit into the above three classes i.e it does not contain instances of hate and offence. \\
    Example: ``@AskPlayStation I can’t get the 14 days free trial please fix I don’t have money for ps plus I need this.''
\end{enumerate}

\begin{table}[]
\begin{tabular}{lllll}
\toprule
\textbf{Language} & \textbf{NONE} & \textbf{HATE} & \textbf{OFFN} & \textbf{PRFN} \\
\midrule
English  & 1852 & 158 & 321 & 1377 \\
German   & 1700 & 146 & 140 & 387 \\
Hindi    & 2116 & 234 & 465 & 148 
\end{tabular}
\caption{Training set label distribution: Task 2}
\label{tab:datafine}
\end{table}

In table \ref{tab:datastats}, we list the data size in number of tweets, and in tables \ref{tab:datacoarse} and \ref{tab:datafine}, we provide the number of instances of different classification labels.

\section{Approach}
In this section, we outline our approach towards solving the task at hand.

\subsection{Preprocessing}
We utilized the python libraries tweet-preprocessor\footnote{\url{https://github.com/s/preprocessor}} and ekphrasis\footnote{\url{https://github.com/cbaziotis/ekphrasis}} for tweet tokenization and hashtag segmentation respectively. For extracting English and German cleaned tweet texts, tweet-preprocessor’s clean functionality was used. For Hindi tweets, we tokenized the tweet text on whitespaces and symbols including colons, commas and semicolons. This was followed by removal of hashtags, smileys, emojis, URLs, mentions, numbers and reserved words (such as @RT which indicates Retweets) to yield the pure Hindi text within the tweet.

\subsection{Feature Engineering}
\label{sec:featEng}
In addition to the cleaned tweet, we utilize tweet-preprocessor to populate certain information fields which can act as features for our classifiers. We include the hashtag text which is segmented into meaningful tokens using the ekphrasis segmenter for the twitter corpus. We also save information such as URLs, name mentions such as `@derCarsti’, quantitative values and smileys. We extract emojis which can be processed in two ways. We initially experimented with the emot\footnote{\url{https://github.com/NeelShah18/emot}} python library to obtain the textual description of a particular emoji. For example, `\Large{\emojismilesweat{}{}}\normalsize' maps to `smiling face with open mouth \& cold sweat' and `\Large{\emojipanda{}{}}\normalsize' maps to `panda'. We later chose to utilize emoji2vec \cite{DBLP:journals/corr/EisnerRABR16} to obtain a semantic vector representing the particular emoji. The motivation behind this is as follows: the text describing the emoji’s attributes might not capture all the pragmatics and the true sense of what the emoji signifies in reality. As a concrete example, consider `\Large{\emojitongue{}{}}\normalsize', the tongue emoji. The textual representation will not showcase the emoji’s association with `joking around, laughter and general goofiness’ which is its real world implication. We expect emoji2vec to capture these kinds of associations.

\subsection{Perspective API Features}
We perform experiments with features extracted from the Perspective API. The API uses machine learning models to estimate various numerical metrics modeling the perceived impact which a post or a comment might have within a conversation. Right now, the Perspective API does not support Hindi natural language text in Devanagari script. Thus, our experiments are on German and English. On German text, the API provides scores which are real numbers between $0$ and $1$ for the following fields: `toxicity', `severe toxicity', `identity attack', `insult' and `profanity and threat'. For English text, in addition to the fields for German, the API provides similar scores for the fields: `sexually explicit', `obscene' and `toxicity fast' (which simply uses a faster model for computing toxicity levels on the back-end).

For both English and German tweets, we extract perspective API scores for all available fields using (a) the complete tweet as is, and (b) the extracted cleaned tweet text excluding emojis, smileys, URLs, mentions, numbers, hashtags and reserved words. Thus, we have 18 features for English tweets and 12 features for German tweets to work with.

We trained multi-layer perceptron classifiers for English and German using a concatenation of these features as the input vector. In addition to these classifiers trained in the monolingual setting, we trained an English-German multilingual classifier using the 12 perspective API features which are common to English and German. The datapoints in the corresponding training sets were randomly shuffled and standardized. The same standardization values were used on the test set during inference. We tried out multiple training settings with different activation functions and optimization techniques. The best results with Perspective features are presented in Section \ref{sec:results}.

\begin{figure}
    \centering
    \includegraphics[scale=0.6]{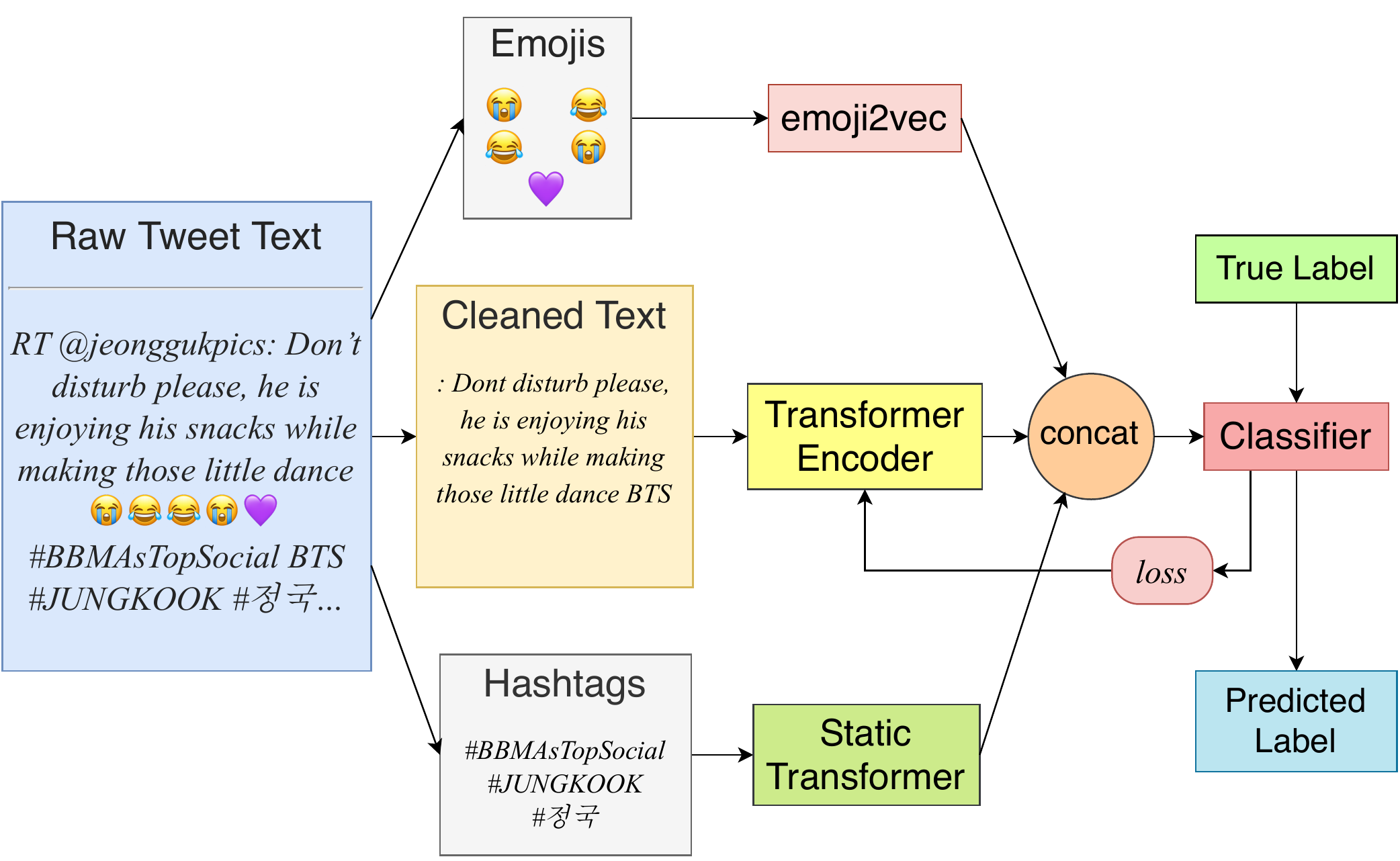}
    \caption{Model Overview}
    \label{fig:pipeline}
\end{figure}

\subsection{Proposed Transformer-based Models}

We leverage Transformer-based \cite{vaswani2017attention} masked language models to generate semantic embeddings for the cleaned tweet text. In addition to the cleaned tweet’s embedding, we generate and utilize semantic vector representations for all the emojis and segmented hashtags available within the tweet. The segmented hash embeddings are generated using the same pre-trained Transformer model such that the text and hashtag embeddings are grounded in the same latent space. emoji2vec is used to create the emojis' semantic embeddings. The Transformer layers encoding the cleaned tweet text are updated during the fine-tuning process on the available training data. For classification, we use the concatenation of the cleaned tweet's embedding with the collective embedding vector for segmented hashtags and emojis.

We are required to encode a list of emojis \& a list of segmented hashtags, both of which can be of variable lengths. Therefore, we average the vector representations of all the individual emojis or segmented hashtags as the case may be, to generate the centralised emoji or hashtag representation. This is simple, intuitive, and earlier work on averaging local word embeddings to generate global sentence embeddings \cite{arora2016simple} has showed that this yields a comprehensive vector representation for sentences. We assume the same to hold true for emojis and hashtags as well.

The concatenated feature-set is then passed to a two layer multi-layer perceptron (MLP). The loss from the classifier is propagated back through the cleaned tweet Transformer encoder during training. We experimented with XLM-RoBERTa (XLMR) \cite{conneau2020unsupervised} as our pre-trained Transformer in various training settings. XLM-RoBERTa has outperformed similar multilingual Transformer models such as mBERT(multilingual BERT) \cite{devlin2018bert} and multilingual-distilBERT \cite{sanh2020distilbert} on various downstream tasks. We therefore chose XLMR as our base Transformer model for the purpose of the shared task. A high level overview of our model flow is shown in figure \ref{fig:pipeline}.



For fine-tuning our XLMR Transformer weights, we perform learning rate scheduling based on the actual computed macro F1-scores on the validation split instead of using the validation loss. As opposed to simply using early-stopping to prevent overfitting, we consider the change in validation performance at the end of each training iteration. If the validation performance goes down across an iteration, we trace back to the previous model weights and scale down our learning rate. Training stops when the learning rate reaches a very small value $\epsilon$\footnote{Set to 1e-12 in our experiments.}. Although expensive, this form of scheduling ensures that we maximize our Macro F1-score on the validation split. For further details on specific implementation nuances and choice of hyperparameters, refer to Section \ref{sec:expt-dets}.

\section{Results}
\label{sec:results}
In this section, we provide quantitative performance evaluations of our approaches on the provided testing-set, the evaluation metric used throughout being the macro F1-score. 



\begin{table}[]
\centering
\begin{tabular*}{\textwidth}{c@{\extracolsep{\fill}} cccccc}
\toprule
 & \multirow{2}{*}{\textbf{Activation}} & \multirow{2}{*}{\textbf{Optimization}} & \multicolumn{2}{c}{\textbf{English}} & \multicolumn{2}{c}{\textbf{German}} \\ \cmidrule{4-5} \cmidrule{6-7}
 & & & Task 1 & Task 2 & Task 1 & Task 2 \\ \midrule
\multirow{2}{*}{monolingual}  & identity & adam (early-stop) & \textbf{89.68} & \textbf{53.90} & 75.40 & 41.84 \\ 
& tanh & adam (early-stop) & 88.93 & 47.07 & \textbf{79.25} & 43.00 \\ \midrule
\multirow{4}{*}{multilingual} & identity & sgd (adaptive LR) & 88.82 & 47.02 & 72.89 & 38.86 \\
& identity & adam (early-stop) & 88.44 & 46.00 & 72.63 & 42.83 \\
& tanh & sgd (adaptive LR) & 87.69 & 44.86 & 75.38 & 38.80 \\
& tanh & adam (early-stop) & 87.95 & 46.03 & 76.68 & \textbf{46.40} \\ 
 
\end{tabular*}
\caption{Perspective API Experiments $($Best results highlighted in bold$)$}
\label{tab:perspective}
\end{table}

In table \ref{tab:perspective}, we present our study on usage of Perspective API features with a multi-layer perceptron classifier for English and German tasks. We notice that these features are able to provide respectable results on the hate and offensive content detection but cannot compete with the Transformer-based models when fine-grained classification is required. In the monolingual mode, our exhaustive grid search showed that the use of identity activation for English and tanh activation for German are the most effective MLP hidden layer activation settings. Table \ref{tab:perspective} lists the best activation functions and optimization techniques for particular $($task, language$)$ pairs. We observe that German Task 2 benefits from the multilingual mode and we attribute this to the additional data from the English training examples which allow the model to generalize better. However, a drop in the English results is witnessed which might be due to the reduction in the number of available features.

\begin{table}[]
\centering
\begin{tabular*}{\textwidth}{c@{\extracolsep{\fill}} cccccc}
\toprule
\multirow{2}{*}{\textbf{Model}} & \multicolumn{2}{c}{\textbf{English}} & \multicolumn{2}{c}{\textbf{German}} & \multicolumn{2}{c}{\textbf{Hindi}} \\ \cmidrule{2-3} \cmidrule{4-5} \cmidrule{6-7}
 & Task 1 & Task 2 & Task 1 & Task 2 & Task 1 & Task 2 \\ \midrule
XLMR-freeze-mono & 83.92 & 52.38 & 66.85 & 41.52 & 68.25 & 40.45 \\ 
XLMR-freeze-multi & 82.02 & 51.02 & 68.34 & 48.60 & 66.27 & 41.59 \\ \midrule
XLMR-adaptive & \textbf{90.29} & 59.03 & 81.04 & 52.99 & \textbf{75.40} & 45.87 \\ 
XLMR-tuned & 90.05 & \textbf{60.70} & \textbf{81.87} & \textbf{53.28} & 74.29 & \textbf{49.74} \\ 
\end{tabular*}
\caption{Performance of Proposed Transformer-based Models $($Best results highlighted in bold$)$}
\label{tab:transformers}
\end{table}

In table \ref{tab:transformers}, we present results using our proposed Transformer-based models. We present XLMR-freeze-mono and XLMR-freeze-multi as baselines in which we use the pre-trained XLM-RoBERTa Transformer weights without any fine-tuning\footnote{We used UKPLab's sentence-transformers library's pre-trained model: `xlm-r-100langs-bert-base-nli-mean-tokens' for this task. The model is available at \url{https://github.com/UKPLab/sentence-transformers.}}. Only the classifier head is trained in these models. We train six separate models for the three languages (two tasks per language) and report corresponding results in the monolingual mode. In multilingual mode, we only train two models on the aggregated training data for the two tasks and use that for inference across the three languages.

The models: XLMR-adaptive and XLMR-tuned use our proposed adaptive learn rate scheduling. In XLMR-tuned, the epsilon value of the Adam optimizer was set to 1e-7 as this experimental setting provided gains on the validation split in our hyper-parameter tuning phase. In both of these models, we jointly fine-tune the XLM-RoBERTa Transformer weights and the classifier head in a multilingual setting. Our proposed models significantly outperform baselines with frozen Transformer weights which is both intuitive and expected.

\begin{table*}
  \begin{tabular*}{\textwidth}{c@{\extracolsep{\fill}} ccccccc}
    \toprule
    & \multirow{2}{*}{\textbf{Features}} & \multicolumn{2}{c}{\textbf{English}} & \multicolumn{2}{c}{\textbf{German}} & \multicolumn{2}{c}{\textbf{Hindi}} \\\cmidrule{3-4} \cmidrule{5-6} \cmidrule{7-8}
    & & Task 1 & Task 2 & Task 1 & Task 2 & Task 1 & Task 2 \\ \midrule
\multirow{3}{*}{monolingual}& cleaned text & 83.27 & 49.12 & 69.23 & 39.12 & \textbf{68.45} & 45.18 \\
& cleaned text + emoji & \textbf{83.60} & 49.78 & 68.05 & 40.54 & 68.21 & 45.48 \\ 
& cleaned text + hashtag & 83.17 & \textbf{53.60} & 66.23 & 41.91 & 66.98 & \textbf{50.08} \\ \midrule 
\multirow{3}{*}{multilingual}& cleaned text & 80.47 & 47.88 & 71.07 & 46.66 & 64.84 & 44.39 \\ 
& cleaned text + emoji & 82.73 & 51.90 & \textbf{72.73} & 43.24 & 67.83 & 41.83 \\ 
&cleaned text + hashtag & 81.22 & 50.06 & 68.52 & \textbf{47.31} & 68.19 & 44.71 \\ 
\end{tabular*}
\caption{Feature Selection Study $($Best results highlighted in bold$)$}
\label{tab:monobase}
\end{table*}

Finally, in table \ref{tab:monobase}, we show results for a study on feature selection using pre-trained XLM-RoBERTa as the Transformer architecture for generating text embeddings. Note that our primary models including XLMR-freeze utilize all of the discussed features. Like XLMR-freeze, the Transformer layers are frozen and not fine-tuned during the training process. The table is separated into monolingual and multilingual modes of training. Results are showed using different feature collections, namely, `cleaned tweet text only', `cleaned tweet + hashtags', and `cleaned tweet + emojis' as inputs to the classifier. We observe a performance drop for English and Hindi and a considerable performance gain for German while moving from monolingual to multilingual training settings.


\section{Experimental Details}
\label{sec:expt-dets}
We used Hugging Face's\footnote{\url{https://huggingface.co/}} implementation of XLM-RoBERTa in our proposed architecture. Our architectures using Transformer models with custom classification heads were implemented using pytorch\footnote{\url{https://pytorch.org/}}. We used Adam optimizer for training with an initial learning rate of 2e-5, dropout probability of 0.2 with other hyper-parameters set to their default values. We updated weights based on cross-entropy loss values. For studies with Perspective API Features and experiments where we do not fine-tune the Transformer weights, we used scikit-learn's \cite{scikit-learn} implementation of a multi-layer perceptron and UKPLab's sentence-transformers library \cite{reimers-2020-multilingual-sentence-bert} whenever applicable.

In our Perspective API experiments, we used deep multi-layer perceptrons with 12 and 9 hidden layers for the binary and multi-class classification modes respectively. Across all our experimental settings, we used a batch size of 200 with other hyper-parameter values set to default. We performed an exhaustive grid search for every multi-layer perceptron model varying the activation function, size of hidden layer, optimization algorithm and type of learning rate scheduling. We reported results using the grid search settings which performed the best on a 4-fold cross validation on the training set. Our experimentation code is publicly available at \url{https://github.com/sayarghoshroy/Hate-Speech-Detection}.

\section{Conclusion}

In this paper, we have leveraged the recent advances in large scale Transformer-based language model pre-training to build models for coarse detection and fine-grained classification of hateful and offensive content in social media posts. Our experiments showcase the utility and effectiveness of language models pre-trained with multi-lingual training objectives on a variety of languages. Our studies show the efficacy of Perspective API metrics by using them as standalone features for hate speech detection. Our best model utilized semantic embeddings for cleaned tweet text, emojis, and segmented hashtags as features, and a customized two-layer feed-forward neural network as the classifier. We further conducted a feature selection experiment to view the impact of individual features on the classification performance. We concluded that the usage of hashtags as well as emojis add valuable information to the classification head. We plan to further explore other novel methods of capturing social media text semantics as part of future work.

\bibliography{sample-ceur}

\end{document}